\documentclass[conference]{IEEEtran}
\IEEEoverridecommandlockouts
\usepackage{cite}
\usepackage{amsmath,amssymb,amsfonts}
\usepackage{algorithmic}
\usepackage{graphicx}
\usepackage{textcomp}
\usepackage{xcolor}
\usepackage{breqn}
\usepackage{multirow}
\usepackage{booktabs}
\usepackage{makecell}
\usepackage[T1]{fontenc}
\usepackage{url}
\usepackage{bm}
\usepackage{tikz}
\def\BibTeX{{\rm B\kern-.05em{\sc i\kern-.025em b}\kern-.08em
    T\kern-.1667em\lower.7ex\hbox{E}\kern-.125emX}}
\begin{document}

\title{Textual Explanations for Automated Commentary Driving\\
\thanks{This work was partially supported by a fellowship within the IFI programme of the German Academic Exchange Service (DAAD) and by the EPSRC project RAILS (grant reference: EP/W011344/1).}
}

\author{\IEEEauthorblockN{Marc Alexander Kühn\textsuperscript{\textsection}}
\IEEEauthorblockA{
\textit{UVC Partners}\\
Munich, Germany \\
kuehn@uvcpartners.com}
\and
\IEEEauthorblockN{Daniel Omeiza}
\IEEEauthorblockA{\textit{Oxford Robotics Institute} \\
\textit{Department of Engineering Science}\\
\textit{University of Oxford}\\
Oxford, United Kingdom \\
danielomeiza@robots.ox.ac.uk}
\and
\IEEEauthorblockN{Lars Kunze}
\IEEEauthorblockA{\textit{Oxford Robotics Institute} \\
\textit{Department of Engineering Science}\\
\textit{University of Oxford}\\
Oxford, United Kingdom \\
lars@robots.ox.ac.uk}
}

\maketitle
\begingroup\renewcommand\thefootnote{\textsection}
\footnotetext{During the project, the author was with the Department of Informatics, Technical University of Munich.}
\endgroup

\begin{abstract}
The provision of natural language explanations for the predictions of deep-learning-based vehicle controllers is critical as it enhances transparency and easy audit. In this work, a state-of-the-art (SOTA) prediction and explanation model is thoroughly evaluated and validated (as a benchmark) on the new Sense--Assess--eXplain (SAX). Additionally, we developed a new explainer model that improved over the baseline architecture in two ways: (i) an integration of part of speech prediction and (ii) an introduction of special token penalties. On the BLEU metric, our explanation generation technique outperformed SOTA by a factor of 7.7 when applied on the BDD-X dataset. The description generation technique is also improved by a factor of 1.3. Hence, our work contributes to the realisation of future explainable autonomous vehicles.
\end{abstract}


\newcommand\copyrighttext{
  \footnotesize ©2022 IEEE. Personal use of this material is permitted. Permission from IEEE must be obtained for all other uses, in any current or future media, including reprinting/republishing this material for advertising or promotional purposes, creating new collective works, for resale or redistribution to servers or lists, or
reuse of any copyrighted component of this work in other works.}

\newcommand\copyrightnotice{
\begin{tikzpicture}[remember picture,overlay]
\node[anchor=south,yshift=10pt] at (current page.south) {\fbox{\parbox{\dimexpr\textwidth-\fboxsep-\fboxrule\relax}{\copyrighttext}}};
\end{tikzpicture}%
}
\copyrightnotice

\section{Introduction}
Deep neural networks typically operate as black-boxes without providing adequate insights into their decision-making process. This makes the justification of their usage in high-stakes scenarios difficult \cite{Kim2018}. Explainable models are handy as they provide better insights into the internal workings of the underlying opaque models. While the end-to-end learned model in~\cite{Kim2018} reports state-of-the-art results, it has only been tested on a single dataset that incorporates only post-hoc explanations in free text form. Hence, a thorough evaluation of the language generation model on additional datasets is essential.

\begin{figure}[t]
\includegraphics[width=0.425\textwidth]{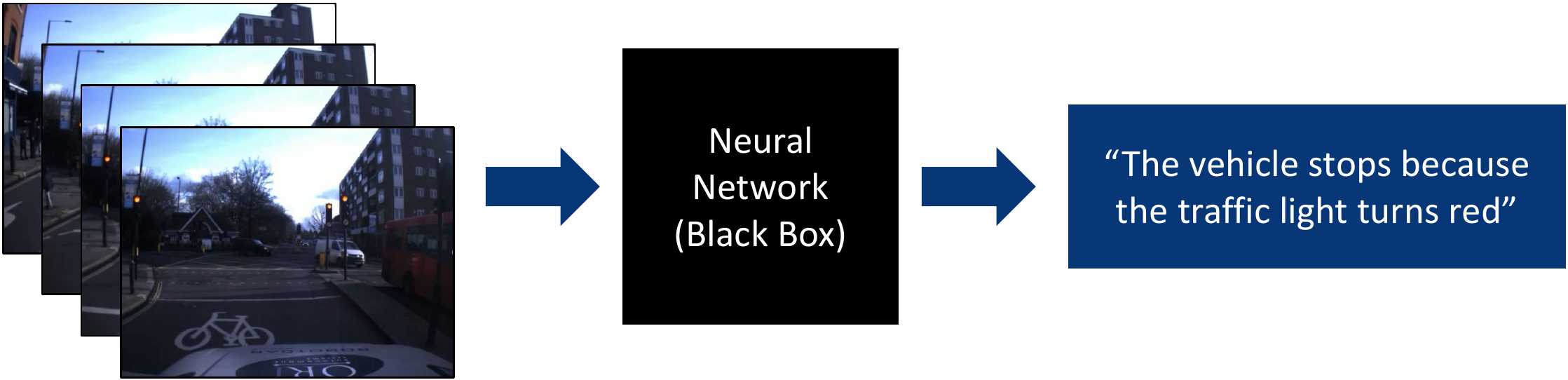}
\centering
\caption{On a high level, the approach takes a series of image frames as input into an opaque neural network and outputs a natural language explanation.}
\label{fig:approach}
\end{figure}

In this paper, we first use a novel \emph{Explanation Driving Dataset}\footnote{Contact authors for access.} which was collected during the Sense--Assess--eXplain (SAX) project~\cite{sax}. The dataset incorporates real-time driver audio commentary into its annotation corpus which allows learning a model on more realistic explanations. Moreover, the natural language annotations were generated and presented in a more structured form which was guided by a multi-label visual annotation process. We evaluated the SOTA model~\cite{Kim2018} on the new dataset (Section~\ref{sec:evaluation}). Also, we implemented and evaluated a new architecture similar to~\cite{Kim2018} but with two modifications; (i) an integration of part of speech prediction and (ii) an introduction of special token penalties. Overall, this paper makes the following contributions:

\begin{enumerate}
  \item an evaluation of the state-of-the-art pipeline on the new SAX data.
  \item integration of part of speech prediction in the language generator.
  \item an improved sentence structure via the inclusion of special token penalties into the loss function.
\end{enumerate}
Fig. \ref{fig:approach} shows the high-level process overview.
Our code is accessible via GitHub\footnote{\url{https://github.com/cognitive-robots/causal-natural-language-explanations}}.


\section{Related Work}
\label{sec:rel_work}
\subsection{Explainable AI for Self-Driving Vehicles}

It is argued that explainable AI in autonomous driving holds many benefits, among which are the enhancement of accountability and facilitation of trust~\cite{arrieta2020explainable,Omeiza_2021}. One commonly used categorisation of explainable AI techniques is based on the explanation provision timing---that is, model intrinsic (inherently interpretable), and post-hoc (built upon a black-box model to explain predictions). Interpretable driver controllers have been proposed in previous works. For example, real-time visual attention maps which  highlight causally influential regions in driving image frames were proposed in~\cite{Li2022}, \cite{Kim2017}. The architecture pipeline consists of a visual attention-based CNN with a subsequent causal filtering technique. Kim et al.~\cite{Kim2018} imported a human-created knowledge corpus consisting of free-text action descriptions and explanations into an attention-based deep learning vehicle controller to provide natural language action descriptions and explanations based on a series of image frames. This approach can also be referred to as an image captioning method. Image captioning is the task to generate a sentence explaining an input image \cite{Mori21}. Most recent methods are based on a CNN- and RNN-based encoder-decoder model like presented by \cite{Feng_2019_CVPR} in an unsupervised set-up, by \cite{Yang_2019_CVPR} incorporating scene graphs or by \cite{Wang_Chen_Hu_2019} using a hierarchical attention network. While those publications use domain-unspecific general image data, \cite{Kim2018} and \cite{Mori21} showed that image captioning methods trained on specific driving data can create textual explanations for driving decisions. In a related work~\cite{Kim2019}, a knowledge corpus consisting of natural language human advice to the driver/AI controller was added. A tree-based approach for automated driving commentary generation has also been explored in~\cite{omeiza2022a} for select driving actions, such as stop, move, and lane changes.

\subsection{Berkeley DeepDrive eXplanation Dataset}
The Berkeley DeepDrive eXplanation (BDD-X) dataset~\cite{Kim2018} consists of videos captured by a dashcam. In addition to the video data, the set also contains timestamped sensor measurements like the vehicle's velocity, course and GPS location. For each scene, the textual annotation consists of an action description and an explanation part with the corresponding timestamps. An example annotation is: \textit{"The car is moving into the right lane because it is safe to do so."}. The annotator writes the description and the explanation part separately in two individual free-text boxes. The dataset consists of over 77 hours of driving scenes within 6,984 videos with an average of 3.8 high-level driving actions. The vocabulary of training action descriptions is 906 words and of explanations 1,668 words. All annotations are conducted post-hoc such that no insights into the driver's real decision-making process are provided.

\subsection{SAX Explanation Driving Dataset}
One key contribution of this work is the evaluation of the state-of-the-art natural language explanation model with the new Sense-Assess-eXplain (SAX) explanation driving dataset. The dataset was obtained during the SAX project~\cite{sax} and it extends existing explanation datasets like BDD-X to provide more structured information through real-time commentary driving. It contains 9.5 hours of driving scenes from London roads with real-time driver audio commentary. The explanation subset contains camera, GPS and CAN-bus data and is visually annotated with the agents' type, action and position. In addition, agents are labelled to indicate how much of an influence they were on the driver's action. This was jointly based on the annotators' judgments and the commentary provided by the driver. Annotators were familiar with the UK road rules. After the completion of the visual annotation, the labels are parsed and agents that had influences are selected and structured to make ground truth explanations for the corresponding ego vehicle's actions. This creates an annotation corpus with standardised sentences and allows to have insights into the real decision-making process of the driver. An example annotation is: \textit{"Car is stopping because pedestrian is crossing on ego's lane"}. Most action sequences are circa 3 seconds long. In total, the dataset consists of 491 driving sequences and the used vocabulary consists of 34 words.

\section{Prediction and Explanation Model}
\label{sec:methodology}

\begin{figure}[b]
\includegraphics[width=0.45\textwidth]{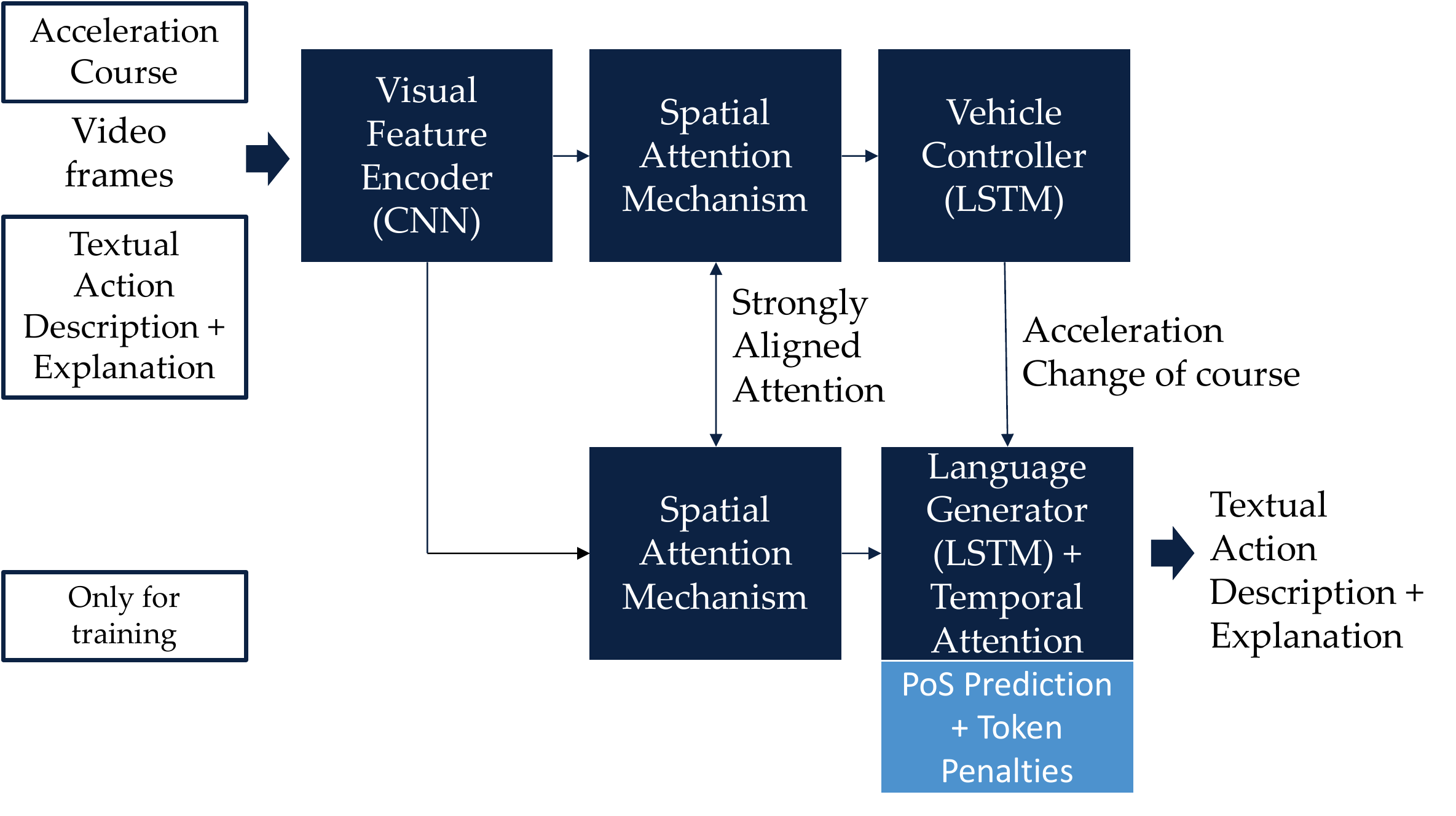}
\centering
\caption{The pipeline consists of a visual feature encoder, a vehicle controller that uses spatial attention and a language generator that uses strongly aligned spatial attention plus temporal attention. We extended the standard model with PoS prediction and special token penalties.}
\label{fig:Model}
\end{figure}

We implemented the SOTA attention-based deep learning model proposed in~\cite{Kim2018}. The model predicts high-level driving actions, such as moving, stopping, changing lanes, and turning. The explanations consist of attention maps accompanied by generated natural language texts. A description of the model is provided in Fig. \ref{fig:Model}. The model is designed to obtain video frames as input and generates natural language action descriptions and explanations as output. In the training, vehicle acceleration and vehicle course values are fed to the model along with corresponding ground truth natural language annotations. Structurally, the pipeline consists of a CNN-based visual feature encoder that extracts visual information which represent high-level object descriptions. Afterwards, an LSTM-based vehicle controller with implemented spatial attention predicts vehicle acceleration and change of course for a series of image frames. In the end, an LSTM-based language generator uses the controller's spatial attention together with temporal attention over multiple video frames. The textual action descriptions and explanations are generated by outputting per-word softmax probabilities.
	
In order to achieve higher-quality textual action descriptions and explanations, we modified the language generator in~\cite{Kim2018} with the aim of generating better sentence grammar. Further, we integrated part of speech prediction in the generator and included special token penalties in the loss function. 

\subsection{Integration of Part of Speech Prediction}

The idea is to include an explicit understanding of part of speech (PoS) into the model. In theory, having constraints on the part of speech could benefit the sentence grammar and improve the similarity score between the generated and ground truth sentences. First, the PoS needs to be added to the natural language annotation corpus for both datasets. For that, each caption is fed into the pre-trained part-of-speech-tagger by \cite{Bird2009} using the universal tagset. The tagger processes sequences of words and adds PoS tags to each word. The used "universal tagset" categorizes adjectives, adpositions, adverbs, conjunctions, determiners, nouns, numerals, particles, pronouns, verbs and punctuation marks. The tokens \textit{<sep>, <START>, <END>} and \textit{<NULL>} are replaced with \textit{";"} tokens such that they get tagged uniformly as \textit{punctuation marks}. 

During training, the deep learning model needs to be adjusted to predict the PoS for each word. This involves the implementation of an additional decoder layer that takes as input the LSTM hidden state $\bm{h}_{k}$ and temporal context vector $\bm{z}_{k}$ and produces part-of-speech predictions $\bm{o}_{k}^{pos}$ for each word according to Equation~\ref{eq:Gen_decoder_POS_1}. \textbf{W} refers to trainable weight matrices. These predictions are an additional input to the subsequent decoder layer that  produces the word predictions $\bm{o}_{k}$. 

\begin{dmath}
\bm{o}_{k}^{pos}=\operatorname{softmax}\left(\left(\mathbf{h}_{k} \cdot \mathbf{W}^{h\_pos}+\mathbf{z}_{k} \mathbf{W}^{z\_pos}\right) \\ \cdot \mathbf{W}^{gen\_out\_pos} \right)
\label{eq:Gen_decoder_POS_1}
\end{dmath}


The loss function of the language generator is additively extended with a new cross entropy loss that takes as input the ground truth one-hot encoded PoS data $p_{k}^{pos}(x)$ together with the generated PoS outputs $\bm{o}_{k}^{pos}$ according to Equation \ref{eq:loss_explainer_pos}. $k$ is an iterator over each word of the sentence and $T$ describes the tagset containing all possible PoS tags.

\begin{equation}
\mathcal{L}^{pos}_{\mathrm{g}}=-\sum_{k} \sum_{x \in T} p_{k}^{pos}\left( x \right) \log\left( \bm{o}_{k}^{pos} \left( x \right) \right) 
\label{eq:loss_explainer_pos}
\end{equation}

The overall generator loss with PoS prediction is constructed out of the new PoS loss $\mathcal{L}^{pos}_{\mathrm{g}}$ weighted by $\lambda_{pos}$ and the original generator loss $\mathcal{L}_{\mathrm{g}}$ weighted by $(1-\lambda_{pos})$.


\subsection{Integration of Special Token Penalties}
The raw caption data used in this project has a unique syntax consisting of a \textit{<START>} token in the beginning, a \textit{<sep>} token between action description and explanation, an \textit{<END>} token in the sentence end and possibly several \textit{<NULL>} tokens as padding towards the maximum sentence length. An example caption looks like this: \textit{"<START> the car accelerates to a constant speed <sep> because the light has turned green <END> <NULL> <NULL> <NULL> <NULL> <NULL> <NULL>"}. To steer the model towards generating the correct syntax, deviations from an ideal sentence structure should be penalised. 

As a first step, each generated \textit{<NULL>} token will be penalised with the penalty $\gamma_{null}$. This should steer the model towards trying to predict a word instead of a \textit{<NULL>} token in case of uncertainty towards the end of the sentence. Therefore, the sentence lengths of generated and ground truth sentences should be further aligned. The penalties are additively included in the model loss function. 
Secondly, each deviation from having the tokens \textit{<START>, <sep>, <END>} exactly once per sentence should be penalised as well. In this process, the mentioned tokens are counted in each generated caption. In the case of $\#token \neq 1$, a penalty $\gamma_{other}$ once per respective token is added to the loss function.
	

\section{Experiments and Evaluation}
\label{sec:evaluation}
\subsection{Training Procedure and Evaluation Set-Up}
For the training of the language generator, an exponential learning rate decay with a decay rate of 0.96 was implemented together with an Adam optimizer \cite{Kingma2014}. For the BDD-X dataset, the decay was applied every 11,600 steps and for the SAX dataset every 3,500 steps. For the BDD-X dataset, a start learning rate of $1e^{-5}$, a batch size of 32 and a dictionary with 1,300 words were used. For the SAX dataset, a start learning rate of $6e^{-7}$, a batch size of 16 and a dictionary with 34 words were used. Furthermore, TensorFlow version 1.15 was used and the graph-level random seed for the default graph was fixed for better reproducibility \cite{Abadi2015}. Moreover, an NVIDIA GeForce RTX 2080 Ti with 11GB GDDR6 memory was used. The datasets were split between training, validation and test sets with ratios of 80\%/10\%/10\% for BDD-X and 75\%/12.5\%/12.5\% for SAX, respectively.

Two quantitative performance metrics were used: METEOR and BLEU. The METEOR metric was implemented following the algorithm in~\cite{Lavie2007} and using the Natural Language Toolkit from~\cite{Bird2009}. METEOR functions by using word-to-word matches between a reference and candidate string which create word alignment. The METEOR score is then calculated based on a parameterized harmonic mean of unigram precision and unigram recall taking also into account a penalty referring to the correct word order. For the BLEU metric, SacreBleu~\cite{Post2018} which wraps the original reference implementation in~\cite{Papineni2002} with additional features to provide comparable corpus-level scores was used. The main idea of BLEU is to compare n-grams of the candidate and reference string by counting the number of position-independent matches. For that, the modified n-gram precision metric is introduced with an additional brevity penalty referring to the sentence length. In addition, we performed a qualitative analysis of selected generated texts. The samples were categorised as good or bad samples according to their assigned METEOR score. The best-performing model parameters from our experiments were used: $\lambda_{pos} = 0.3$, $\gamma_{null}=4$ and $\gamma_{other}=50$. 

\subsection{Standard Model Performance on SAX and BDD-X}

\subsubsection{Quantitative Results}
\begin{table}[pb]
  \caption{This table shows the corpus-level METEOR and BLEU scores as percentages.}\label{tab:Gen_Perf1}
  \centering
  \resizebox{\columnwidth}{!}{%
  \begin{tabular}{l | c c | c c}
    \toprule
    \multirow{2}{*}{ } & 
    	\multicolumn{2}{c|}{METEOR [\%]} & 
    	\multicolumn{2}{c}{BLEU [\%]} \\
    & Description & Explanation & Description & Explanation \\
    \midrule
    BDD-X (70\%) & 45.92 & 10.47 & 14.82 & 0.33 \\
    SAX & 62.82 & 68.30 & 29.57 & 41.11 \\
    \bottomrule
  \end{tabular}
  }
\end{table}

Table~\ref{tab:Gen_Perf1} shows METEOR and BLEU scores as percentage values for our model implementation trained on 70\% of the BDD-X dataset (due to unavailability of the remaining 30\%) and on the SAX dataset. The scores are individually calculated for the description (e.g. "Car is stopping") and explanation part (e.g. "because traffic light is not green on ego's lane") of each generated sentence. 

On the BDD-X dataset, the model generated more accurate phrases for the action description part than for the explanation part: The METEOR score for the description is $4.4\times$ higher while the BLEU score is $44.9\times$ higher. For the SAX dataset, such a significant difference was not observed. The scores for the explanation were slightly higher than the description part. Explanation scores were 41.11\% against 29.57\% (BLEU) and 68.30\% against 62.82\% (METEOR). Overall, the language generator creates significantly higher scoring samples for the SAX dataset than for the BDD-X dataset, especially, for the explanation part.

\subsubsection{Qualitative Analysis}

From the qualitative analysis performed using the BDD-X data, results indicate that no sample with a completely correct sentence was generated. The most common errors were word repetitions and single missing words in the generated sentences. For the good scoring samples, the semantics were captured correctly by the language generator. The image frames had clearer visuals. The bad scoring samples did not contain correct grammar (e.g. \textit{"the car is forward"} or \textit{"a to turn"}) and did not express the correct meaning. The ground truth sentences are longer and more complicated compared to the good-scoring samples. Additionally, the video frames had unclear visuals like sun glare or bad illumination. 

For the SAX dataset, clear differences in quality between action description and explanation were observed: The description part is within the good scoring samples mostly correctly generated. The explanation part is not once perfectly generated. A few samples come close, for example by substituting the word \textit{"ego"} with \textit{"car"}. Although the bad sentences have mostly valid grammar, semantics are wrongly captured. 

\subsubsection{Discussion}
One reason why the BDD-X trained model scores higher on the description than the explanation part is because describing a scenario is semantically easier than finding a causal explanation. Another possible reason is based on sentence grammar: The free text annotation process of the BDD-X set results in largely varying samples, therefore, learning the sentence syntax and length could be especially difficult for the model. Probably, this has a stronger impact on the explanation part due to more variant expressions for post-hoc explanations than for descriptions.

The model trained on the SAX dataset generated higher scores because the ground truth sentences were annotated in a more structured way. This results in less variety of used words and sentence lengths which allows the model to generate high-scoring samples easily. One reason for this might be that the generation of the right amount of \textit{NULL} tokens towards the end of the sentence was comparably easy using the uniform SAX samples and positively influences the explanation scores. Another reason might be due to more realistic ground truth explanations, because of the incorporated real-time driver commentary of the SAX set. This could increase data quality and therefore ease the model's learning process.

On the qualitative analysis, one can conclude that further improvements in language generation are needed. In the case of the BDD-X dataset, especially word repetitions, grammar and sentence lengths need to be improved. The generation of shorter sentences is probably due to the tendency of the network to generate \textit{<NULL>} tokens towards the end of the sentence instead of trying to predict the correct word as a result of uncertainty. For the SAX dataset, grammar is mostly correctly generated, but wrong semantics are often expressed. A reason for that could be the small dataset size nature of SAX compared to the BDD-X set.


\subsection{Evaluation of Language Generator Modifications}

\subsubsection{Quantitative Results}
Table~\ref{tab:Gen_Var_BDDX} shows the METEOR and BLEU scores for the new model variants compared to the standard model using the BDD-X dataset. For both description and explanation generation, the model variations generated higher-scoring samples. The best explanations were generated using the combination of PoS prediction and special token penalties. The improvement is significant as it yields METEOR scores that were $1.5\times$ better and BLEU scores that were $7.7\times$ higher. Although, absolute scores were still significantly lower compared to the description scores. The best descriptions were generated using only token penalties. Here, the improvement is less significant: The METEOR score improves only slightly by $1.6\%$ whereas the BLEU score improves more by $5.1\%$ which equals $1.3\times$ better performance.

Table~\ref{tab:Gen_Var_SAX} shows the METEOR and BLEU scores for the model variants using the SAX dataset. Here, a more significant improvement can be seen for description generation: The combination of PoS prediction and token penalties performs best with a $10.9\%$ higher METEOR score and $5,8\%$ higher BLEU score. Regarding explanation generation, the best-performing model consists only of PoS prediction with marginal improvement: $0.8\%$ higher according to METEOR and $0.7\%$ higher according to BLEU. Both models that included special token penalties generated worse scoring explanations compared to the standard model on SAX: 5.9-8.2\% lower METEOR scores and 3.3-8.1\% lower BLEU scores. 

\begin{table}[tp]
  \caption{This table shows the corpus-level METEOR and BLEU scores as percentages for the variations of the language generator on the BDD-X dataset.}\label{tab:Gen_Var_BDDX}
  \centering
  \resizebox{\columnwidth}{!}{%
  \begin{tabular}{l | c c | c c}
    \toprule
    \multirow{2}{*}{ } & 
    	\multicolumn{2}{c|}{METEOR [\%]} & 
    	\multicolumn{2}{c}{BLEU [\%]} \\
    & Description & Explanation & Description & Explanation \\
    \midrule
    Standard Model & 45.92 & 10.47 & 14.82 & 0.33\\
    Standard + PoS Prediction & 46.06 & 12.49 & 17.54 & 0.42 \\
    Standard + Token Penalties & \textbf{47.56} & 12.47 & \textbf{19.95} & 1.12 \\
    \makecell[l]{Standard + PoS Predicton \\+ Token Penalties} & 44.64 & \textbf{15.95} & 17.44 & \textbf{2.53} \\
    \bottomrule
  \end{tabular}
  }
\end{table}

\begin{table}[bp]
  \caption{This table shows the corpus-level METEOR and BLEU scores as percentages for the variations of the language generator on the SAX dataset.}\label{tab:Gen_Var_SAX}
  \centering
  \resizebox{\columnwidth}{!}{%
  \begin{tabular}{l | c c | c c}
    \toprule
    \multirow{2}{*}{ } & 
    	\multicolumn{2}{c|}{METEOR [\%]} & 
    	\multicolumn{2}{c}{BLEU [\%]} \\
    & Description & Explanation & Description & Explanation \\
    \midrule
    Standard Model & 62.82 & 68.30 & 29.57 & 41.11\\
    Standard + PoS Prediction & 66.27 & \textbf{69.10} & 29.81 & \textbf{41.84} \\
    Standard + Token Penalties & 64.69 & 60.12 & 34.75 & 32.95 \\
    \makecell[l]{Standard + PoS Predicton \\+ Token Penalties} & \textbf{73.76} & 62.40 & \textbf{35.37} & 37.84 \\
    \bottomrule
  \end{tabular}
  }
\end{table}

Table~\ref{tab:Gen_Var_Weights_BDDX} shows variations of $\lambda_{pos}$ and $\gamma_{null}$ and their impact on the METEOR and BLEU scores. Varying the part of speech weight $\lambda_{pos}$ does not significantly change the performance. The differences between $\lambda_{pos}=0.15$ and $\lambda_{pos}=0.3$ according to METEOR (0.2-0.5\%) and BLEU (0.1-0.7\%) are minimal. Varying the NULL penalty $\gamma_{null}$ led to more significant performance differences: Here, a lower penalty led to higher scoring sample generation - especially for the description part: 47.23\% (METEOR) and 20.38\% (BLEU) for $\gamma_{null}=4$ compared to 44.15\% (METEOR) and 17.09\% (BLEU) for $\gamma_{null}=12$.

\begin{table}[tp]
  \caption{This table shows the corpus-level METEOR and BLEU scores as percentages on the BDD-X dataset for different $\lambda_{pos}$ and $\gamma_{null}$. For the shown results: $\gamma_{other}=0$.}\label{tab:Gen_Var_Weights_BDDX}
  \centering
  \resizebox{\columnwidth}{!}{%
  \begin{tabular}{l | c c | c c}
    \toprule
    \multirow{2}{*}{ } & 
    	\multicolumn{2}{c|}{METEOR [\%]} & 
    	\multicolumn{2}{c}{BLEU [\%]} \\
    & Description & Explanation & Description & Explanation \\
    \midrule
    $\lambda_{pos}=0$ +  $\gamma_{null}=0$ & 45.92 & 10.47 & 14.82 & 0.33\\
    $\lambda_{pos}=0.15$ +  $\gamma_{null}=0$ & 45.60 & 12.69 & 16.76 & 0.52 \\
    $\lambda_{pos}=0.3$ +  $\gamma_{null}=0$ & 46.06 & 12.49 & 17.54 & 0.42 \\
    $\lambda_{pos}=0$ +  $\gamma_{null}=4$ & 47.23 & 11.57 & 20.38 & 0.54 \\
    $\lambda_{pos}=0$ +  $\gamma_{null}=12$ & 44.15 & 10.64 & 17.09 & 0.43 \\
    \bottomrule
  \end{tabular}
  }
\end{table}

\subsubsection{Qualitative Analysis}
Qualitatively, it can be seen that the good-scoring samples capture the correct semantics. Additionally, the sentences were almost identical to the ground truth. For the bad scoring samples, the explanation part represents no complete and syntactically correct sentence. The description parts had valid syntax, but the wrong meaning. On top of that, the ground truth explanations had complex sentence structures.

\begin{figure}[t!]
\centering
\includegraphics[width=0.45\textwidth]{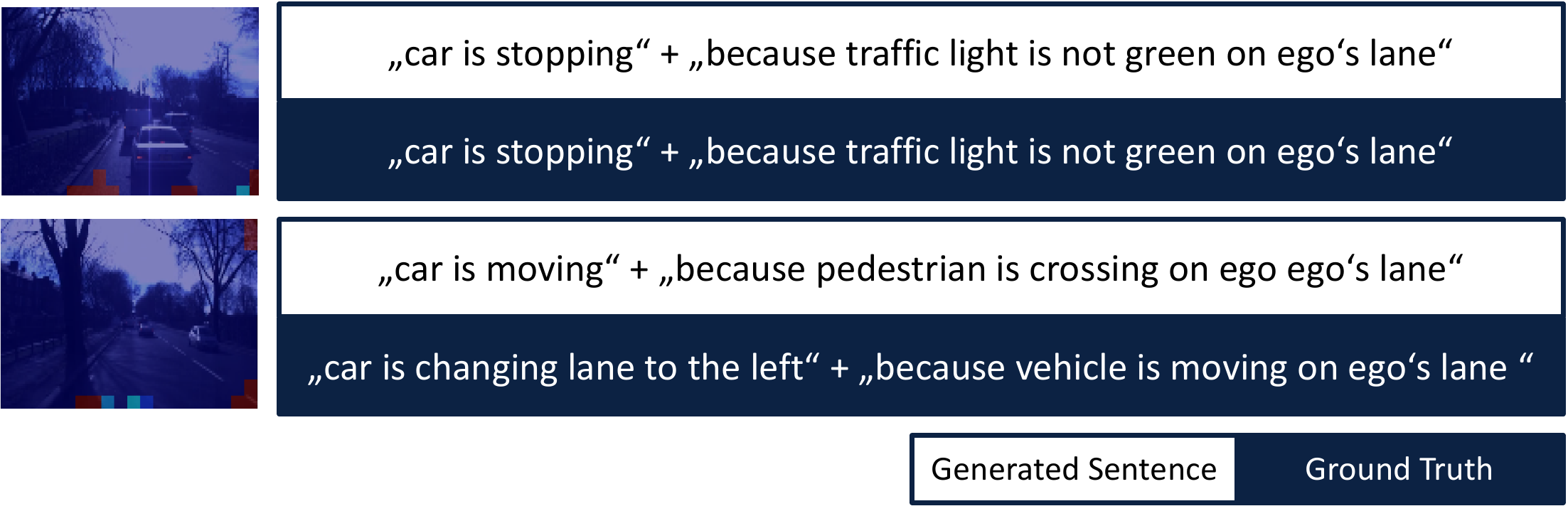}
\caption{This figure shows a good (top) and a bad (bottom) scoring generated sample using the model with PoS prediction trained on the SAX dataset. A corresponding video frame with the spatial attention map as an overlay is shown next to each sample.}
\label{fig:SAX_samples_improved}
\end{figure}

For the SAX dataset, Fig. \ref{fig:SAX_samples_improved} shows a good and a bad scoring sample. For the good, the action description and explanation are correctly generated. The bad sentence has description parts with correct grammar, whereas the explanation sentences had few cases of word repetitions, with the semantics mostly captured wrongly.

\begin{figure}[t]
\centering
\includegraphics[width=0.325\textwidth]{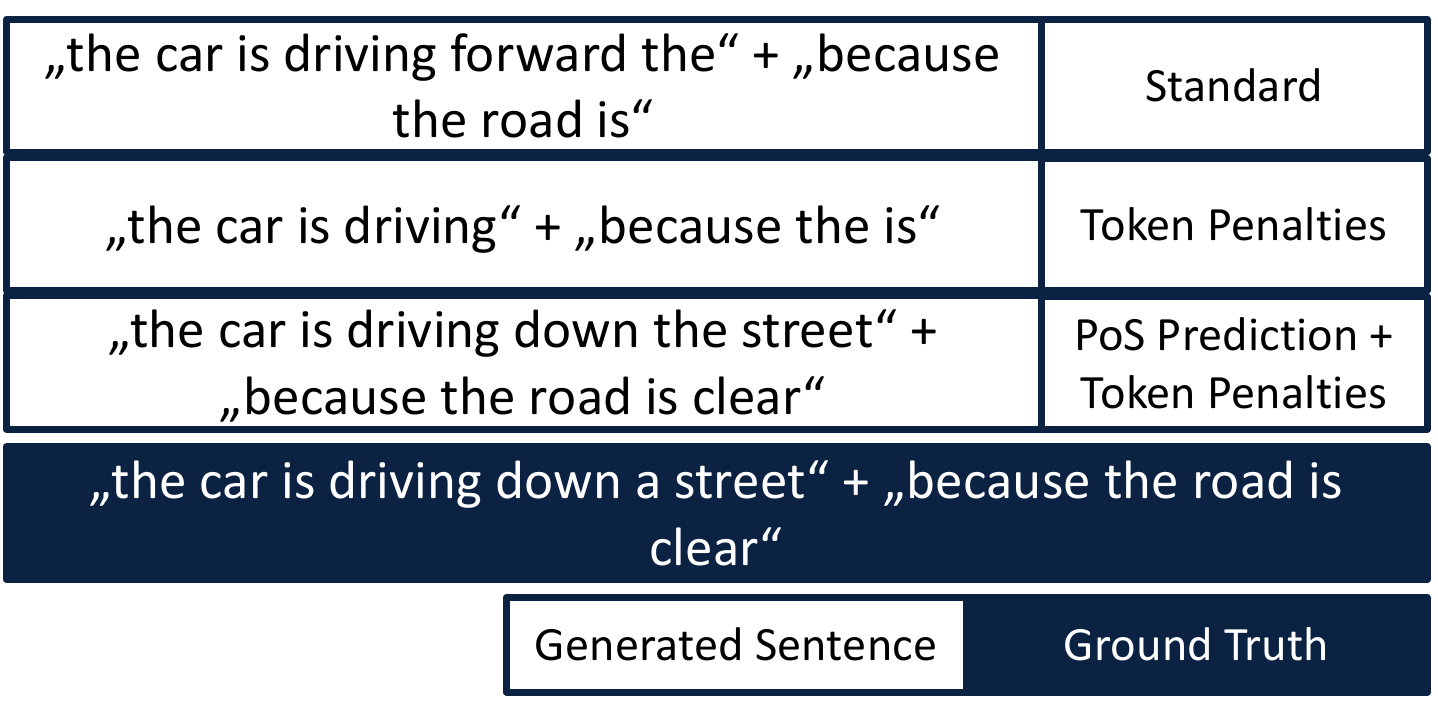}
\includegraphics[width=0.325\textwidth]{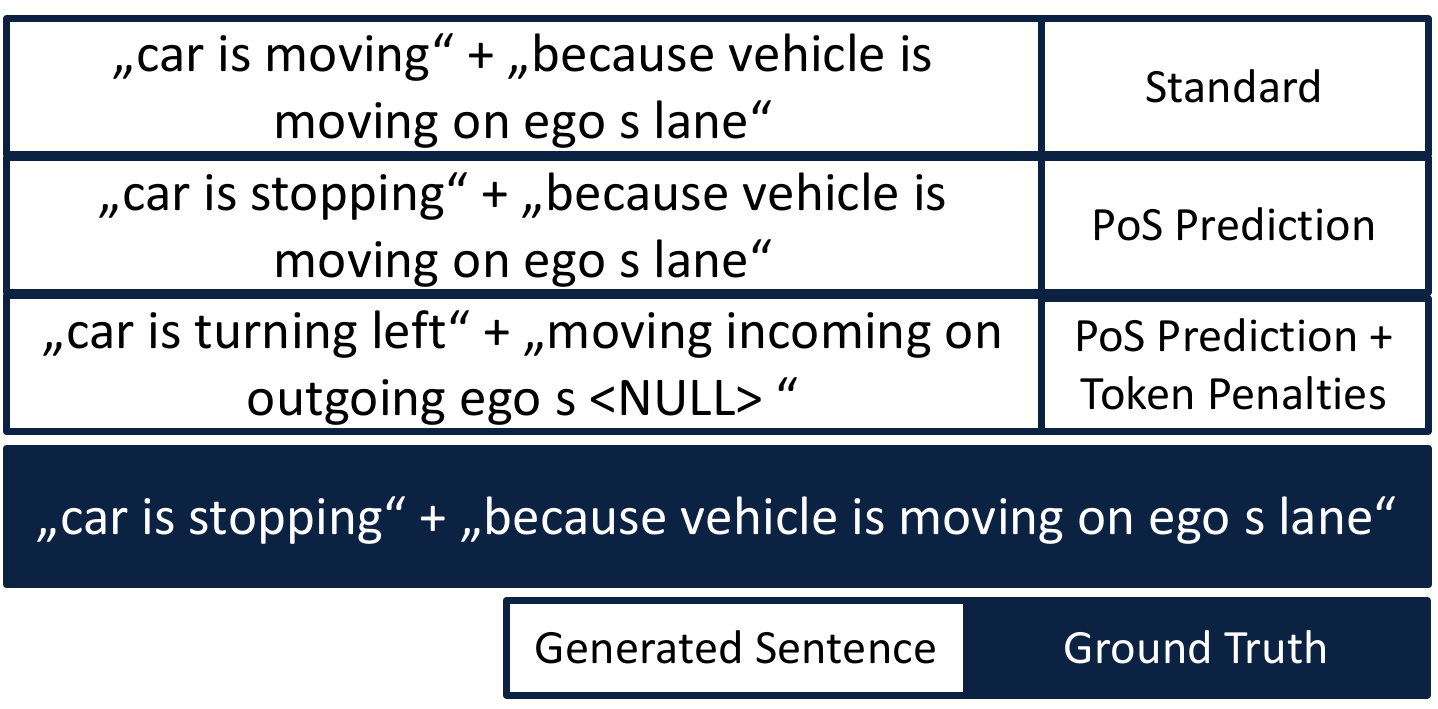}
\caption{This figure shows one sample of the BDD-X (top) and of the SAX dataset (bottom) with its ground truth annotation and three generated texts by the standard model, the model with special token penalties (BDD-X) / with PoS prediction (SAX) and the combination model with PoS prediction and token penalties.}
\label{fig:variations}
\end{figure}

In Fig. \ref{fig:variations} one sample of the BDD-X dataset is shown (top) as ground truth together with three generated sentences according to the standard model, the special token penalty model and the combination model of PoS prediction and token penalties. These two variations were chosen, because they were the quantitatively best-performing ones. In the standard model's generation, the last word of each sentence part is missing. Additionally, instead of the term \textit{"driving down"}, the term \textit{"driving forward"} with a similar meaning was generated. In contrast, the combination model generates a nearly perfect sentence. The model using only token penalties has the worst qualitative performance. 

In Fig. \ref{fig:variations} equivalently one sample of the SAX dataset is shown (bottom) with its ground truth annotation. Here, the generated sentences are produced by the standard model, the model with PoS prediction and the combination model. Again, these two variations were chosen, because they were the quantitatively best-performing ones. In the standard models' sentence, the explanation was correctly generated, but the description portrays the wrong meaning, even though the grammar was correct. The PoS prediction model achieves to improve the description by capturing the correct meaning and therefore generating the correct complete sentence. The combination model consisting of PoS prediction and token penalties generated a faulty sentence: Wrong semantics for the description and explanation and wrong sentence structure for the explanation part. 

\subsubsection{Discussion}
On the BDD-X dataset, the model adjustments led to a relevant performance increase. Especially, the token penalties had a significant impact according to the BLEU scores. The highest performance increase can be observed for the explanation part with the combination of PoS prediction and token penalties. A possible reason could be that the quality of the explanation part usually is dependent on the correct prediction of the separator token and the correct amount of NULL tokens in the end. Therefore, the explanation generation is strongly influenced by the token penalties.

For the SAX dataset, only the description part could be significantly improved. Again, the combination of PoS prediction and token penalties yielded the best performance. For the explanation part, token penalties had a negative influence. This could be due to the uniform annotation corpus of the dataset that makes it already manageable to learn the correct syntax and sentence lengths with the original model. Adding additional NULL penalties might have led to predicting wrong random words instead of the correct NULL tokens. 
	
Testing different strengths of $\lambda_{pos}$ and $\gamma_{null}$ led to the conclusion that the weight of the PoS loss $\lambda_{pos}$ has no significant influence on the model performance, but the strength of the null penalty $\gamma_{null}$ has. An optimal $\gamma_{null}$ value exists. 

Qualitatively, the proposed modifications also increased the performance. The sentence structure was mostly correctly generated and the remaining challenges refer to the model capturing the wrong semantics of a scene. This is especially true for examples with complex ground truth captions. The performance of the individual models varies depending on the used dataset - probably due to different complex ground truth annotations. On an individual sample level, the qualitative assessment of the generated sample is not always aligned with the overall quantitative score on the complete dataset as can be seen in Fig.~\ref{fig:variations}.

\section{Conclusion}
\label{sec:conclusion}
This work validates the performance of the SOTA prediction and explanation pipeline for high-level driving actions on the new SAX dataset with improved performance. The unique features of the SAX dataset, e.g., uniformly structured annotation corpus and real-time driver commentary allow the model to score higher and generate sentences with better sentence grammar compared to the BDD-X set. To further address the faulty sentence grammar in many BDD-X-based generated samples, we modify the language generator by adding part of speech prediction and special token penalties in the loss function. This resulted in a  significant improvement of the explanation generation by a factor of $7.7$ and $1.5$ with the BLEU and METEOR metrics, respectively. The description generation was also improved by a factor of $1.3$. Our modified pipeline outperforms the SOTA end-to-end model.

Nevertheless, our work has limitations: First, the text evaluation metrics BLEU and METEOR mainly rely on sentence similarity and, thus, are limited in comparing semantics. In future work, we would validate our results using human evaluation. Second, optimal weight and penalty values would benefit from a more rigorous estimation. Third, the model's ability to learn correct semantics could be further improved by using a higher performance model for feature processing, e.g., a completely attention-based transformer~\cite{Vaswani2017} or using more sophisticated image representations outputted by an advanced encoder like CLIP~\cite{Radford2021}.

\bibliographystyle{ieeetr}
\small{
\bibliography{mybib} 
}
\end{document}